\documentclass[conference]{IEEEtran}
\IEEEoverridecommandlockouts
\usepackage{cite}
\usepackage{amsmath,amssymb,amsfonts}
\usepackage{algorithmic}
\usepackage{graphicx}
\usepackage{textcomp}
\usepackage{xcolor}
\usepackage{float}
\usepackage{rotating}
\usepackage{ upgreek }
\usepackage{mathtools}

\usepackage{comment}

\newsavebox\mybox

\def\BibTeX{{\rm B\kern-.05em{\sc i\kern-.025em b}\kern-.08em
    T\kern-.1667em\lower.7ex\hbox{E}\kern-.125emX}}
\begin{document}

\title{Deep Sequence Modeling for Anomalous ISP Traffic Prediction \\
}

\author{\IEEEauthorblockN{Sajal Saha, Anwar Haque, and\ *Greg Sidebottom}
\IEEEauthorblockA{\textit{Department of Computer Science} \\
\textit{University of Western Ontario, London, ON, Canada}\\
\textit{*Juniper Networks, Kanata, ON, Canada}\\
Email:\{ssaha59, ahaque32\}@uwo.ca, *gsidebot@juniper.net}
}

\maketitle

\begin{abstract}
Internet traffic in the real world is susceptible to various external and internal factors which may abruptly change the normal traffic flow. Those unexpected changes are considered outliers in traffic. However, deep sequence models have been used to predict complex IP traffic, but their comparative performance for anomalous traffic has not been studied extensively. In this paper, we investigated and evaluated the performance of different deep sequence models for anomalous traffic prediction. Several deep sequences models were implemented to predict real traffic without and with outliers and show the significance of outlier detection in real-world traffic prediction. First, two different outlier detection techniques, such as the Three-Sigma rule and Isolation Forest, were applied to identify the anomaly. Second, we adjusted those abnormal data points using the Backward Filling technique before training the model. Finally, the performance of different models was compared for abnormal and adjusted traffic. LSTM\_Encoder\_Decoder (LSTM\_En\_De) is the best prediction model in our experiment, reducing the deviation between actual and predicted traffic by more than 11\% after adjusting the outliers. All other models, including Recurrent Neural Network (RNN), Long Short-Term Memory (LSTM), LSTM\_En\_De with Attention layer (LSTM\_En\_De\_Atn), Gated Recurrent Unit (GRU), show better prediction after replacing the outliers and decreasing prediction error by more than 29\%, 24\%, 19\%, and 10\% respectively. Our experimental results indicate that the outliers in the data can significantly impact the quality of the prediction. Thus, outlier detection and mitigation assist the deep sequence model in learning the general trend and making better predictions.
\end{abstract}


\section{Introduction}
Internet traffic prediction based on historical data is essential for effective design and optimal usages of computer networks. Traffic prediction and forecasting directly impact certain domains such as optimized routing, dynamic bandwidth allocation, admissibility control, etc. It also has usefulness in planning and managing a computer network, particularly in data centers \cite{hu2015coarse}. A better traffic forecasting model can help network providers manage the network proactively to ensure better Quality of Service (QoS). Moreover, anomalies and security attacks such as DoS or SPAM in the internet traffic can also be identified utilizing the traffic prediction mechanism \cite{krishnamurthy2003sketch}.
The internet traffic data can be captured in the form of time series data using different network flow collection protocols such as NetFlow \cite{rossi2010fine}. Time series is chronologically ordered data collected at a discrete-time interval. The traffic forecasting model accepts traffic data in a time series format where each time point represents traffic volume at that specific moment. Time Series Forecasting (TSF) deals with different tools and techniques to analyze the historical data to predict the future \cite{wheelwright1998forecasting}. 

The time series prediction model can be divided into two board categories: the linear and the non-linear models \cite{terui2002combined}. Several statistical models such as AutoRegressive Integrated Moving Average (ARIMA) \cite{box2015time}, Holt-Winter \cite{wheelwright1998forecasting} are capable of learning the linear component of the time series data and predicting the next value. On the other hand, the non-linear element of the time-series data can be captured by Neural Network (NN) based model such as multi-resolution Finite-Impulse-Response (FIR) model \cite{ alarcon2006multiresolution}, Genetic Algorithm and Radial Based Function Network (GA-RBF) \cite{ wang2008internet } etc. Also, there are some non-linear statistical models e.g. Threshold AutoRegressive (TAR) \cite{ ricky2005threshold}, and Exponential AutoRegressive (ExpAR) \cite{terui2002combined} for handling the non-linear part in time series data. Recently, deep learning models have been used extensively for accurate and efficient prediction in different domains, e.g., predicting the stock market, finance, foreign exchanges, weather forecast, etc. Among various deep neural networks, Recurrent Neural Network (RNN) and its variants such as Long Short Time Memory (LSTM), Bi-directional LSTM, and Gated Recurrent Unit (GRU) received attention due to the capability of sequence analysis. Other deep learning models such as Convolutional Neural Network (CNN), Stacked Denoising Autoencoders (SDA), etc., can also analyze the time series data. 
Real-world IP network traffic is susceptible to various external and internal factors which may abruptly change the normal traffic flow. Internal factors are related to ISP companies, such as introducing new services, traffic migration, speed up-gradation, etc. In contrast, external factors are related to external ISP events and conditions such as new internet applications, regional economic factors, seasonal effects, etc. However, the unexpected suspicious changes, differing significantly from most data points, are prevalent in real IP traffic known as outliers or anomalies. Those abnormal data points in traffic flow might impact learning the general trend in data. As a result, the prediction model might produce wrong inferences in the future, considering them as normal behavior. Therefore, it is essential to identify and handle the anomalies/outlier in the internet traffic before applying any prediction model. In addition, it might help the prediction model to increase the generalization capability. Deep Learning Models have been extensively used to predict the complex real IP traffic. However, the comparative performance analysis among deep sequence models such as RNN and their variants in predicting anomalous internet traffic has not been studied considerably. This work extensively analyzes the real-world IP traffic to identify the abnormal data points and mitigate them before feeding into the deep learning model. We focus on the performance evaluation of deep sequence models in IP traffic prediction with and without the outlier. Our experimental results show better accuracy when the deep sequence models are trained using anomaly-adjusted data. The main contributions of this work are:

\begin{itemize}
    \item Outlier or anomaly detection using the standard statistical procedure and unsupervised learning method before using them to design our predictive model. Also, the mitigation process of the outliers in the data has been discussed. 
   \item Comparative performance analysis of different sequence modeling techniques such RNN, LSTM, LSTM Encoder\_Decoder, LSTM Encoder\_Decoder with Attention layer, and GRU without outlier's treatments.
  \item Comparative performance analysis of the sequence modeling techniques after adjusting the outliers in the data.  
\end{itemize}

This paper is organized as follows. Section \ref{Literature Review} describes the literature review of current traffic prediction using machine learning models. Section  \ref{Methodology} presents the methodology, including dataset description, deep learning models explanation, anomaly identification process, and experiment details. Section  \ref{Results and Discussion} summarizes the different deep learning methods' performance and draws a comparative picture among prediction models with and without outliers in the dataset. Finally, section  \ref{Conclusion} concludes our paper and sheds light on future research directions.

\section{Literature Review}
\label{Literature Review}
P. Cortez et al.\cite{1} proposed three different forecasting methods to predict the volume of internet traffic in TCP/IP-based networks. They investigated both the neural network model and statistical model in their experiment. The proposed novel neural ensemble method performs better in two different time scales, such as 5 minute and hourly forecasting, while the Holt-Winter outperforms daily forecasting. They used the linear interpolation technique to replace the missing data. L. Miguel et al.\cite{2}, compared the performance of the Artificial Neural Network (ANN) model and a statistical model, Holt-Winter, in traffic volume forecasting. The proposed ensemble of Time Lagged Feed-Forward Network (TLFN) explicitly handled the temporal data by incorporating a short-term memory in the input layer of the ANN model.

Furthermore, the sliding window technique has been applied to make the time series data compatible for supervised learning. They also proposed a flow collection methodology for collecting traffic information in a time-series format using the NetFlow protocol. T. P.
Oliveira et al.\cite{3}, compared traffic prediction performance between two different ANN models such as Multi-Layer Perceptron (MLP) and Stacked Autoencoder (SAE). They used two hidden layers of MLP with 60 and 40 neurons, respectively, while they found the best result for four hidden layers SAE with 80, 60, 60, and 40 neurons. They trained their model for 1000 epochs in both MLP and SAE, although the SAE training was divided into two-stage as the unsupervised pre-training for 900 epochs and 100 epochs supervised training.  R. Alfred et al. \cite{4} identified a few drawbacks such as slow convergence, a long training time, and easy to fall into a local minimum of Back Propagation Neural Network (BPNN) in predicting time series data. They proposed a modified version of BPNN (GABPNN), where the initial model weights and threshold were optimized using a Genetic Algorithm (GA). The model has been trained on different configurations to figure out the best-performing hyper-parameters. The overall performance of the GA-based BPNN was significantly better than the BPNN.

C.W. Huang et al. \cite{5} investigated three state-art-of deep learning models in predicting the mobile traffic data. The geographical and temporal properties of the time series data have been extracted using CNN and RNN model, respectively. Their experiment proposes a hybrid model combining CNN and RNN, outperforming other deep and non-deep learning models. Their investigation considered a wide range of parameter settings to identify the best-performing model. W. Wang et al. \cite{6} proposed a novel traffic prediction model called SDAPM based on a stacked denoising autoencoder prediction model (SDA). The SDAPM can extract the generic attributes from the traffic flow. Their model is fine-tuned using a different combination of related hypermeters such as number of hidden layers, number of neurons in the hidden layer, learning rate, etc. The Theano framework \cite{2016arXiv160502688short} has been used as an implementation backend. Finally, R. Madan and P.S. Mangipudi \cite{7} proposed an ensemble traffic prediction model combining the statistical model ARIMA (Auto Regressive Integrated Moving Averages) and the deep learning model RNN (Recurrent Neural Network). The Discrete Wavelet Transform (DWT) transformation technique has been applied to separate the liner and non-linear component from the original time series data. The linear and non-linear parts were analyzed using two different prediction models, ARIMA and RNN, respectively, and combined to predict the final value. Their proposed ensemble model shows better prediction than the individual prediction model.

The deep learning model for predicting internet traffic is getting attention due to the erratic nature of real-world traffic data. Commercial data can have outliers due to unexpected external or internal events, especially in the internet world. Several research works focus on modeling complex deep neural networks or comparing their performance with statistical models. However, we found a lack of investigation in the performance comparison of deep sequence modeling technique such as RNN and their variation in traffic prediction. Also, outlier management in internet traffic prediction using a deep learning model has not been studied extensively. In this work, we examined the performance of deep sequence modeling in real-world internet traffic prediction. Moreover, a statistical and machine learning analysis was also conducted to handle the outlier in our time series data, improving the overall forecast.
\section{Methodology}
\label{Methodology}
In this section, we first introduce the real IP traffic dataset and the proprocessing steps used in our experiment in subsection \ref{data_preprocessing}. Then, we describe the anomaly detection and mitigation techniques to clean our traffic data in subsection \ref{anomaly_detection}. After that, we explain deep learning model background and cross-validation technique in time-series data in subsection \ref{model} and \ref{cross_validation} respectively. The model performance evaluation metrics are described in subsection \ref{metric}. Finally, we summarize the configuration of our experimental environment in subsection \ref{software}.

\subsection{Dataset and Preprocessing Steps}
\label{data_preprocessing}
Real internet traffic telemetry on several high-speed interfaces has been used for this experiment. The data are collected every five minutes for a recent thirty days time period. There are 8563 data samples in our dataset consisting of 29 days of complete data (288 data instances per day), while the last one-day data is incomplete. We considered Only the timestamp (GMT) and traffic data (bit per second) from the original data file in JSON (JavaScript Object Notation) format, and all other information is discarded. Ultimately, a total of 29 days of data were considered for developing our prediction model. The unit of our traffic data has been changed from bps (bit per second) to Gbps (gigabit per second) as the original value is large for feeding into our statistical model.
\subsubsection{Handling Missing Value} 
The last day data from our dataset was removed in our experiment as its networks trace was collected partially on that day. There are 29 missing values in our dataset, which are replaced using the forward filling technique. The previous valid data instance has been used to replace the missing value in our traffic data. There are other methods, such as linear interpolation, quadratic interpolation, backward filling, etc., to handle the missing data in time series analysis. We found the forward filling technique useful for our dataset. 
\subsubsection{ACF/PACF Analysis}
We need to convert our traffic data from time-series format to an array of features and the corresponding target for the supervised prediction model. The Autocorrelation Function (ACF) and Partial Autocorrelation (PACF) analysis help us identify the significant number of past observations on which our next value is highly dependent. Moreover, the significant lag from the ACF and PACF plot can be used as a number of features in the data conversion process where the next value after the lag will be the corresponding target. Hence this analysis is essential to convert our internet traffic data compatible for supervised learning.
\subsubsection{Data Windowing}
\begin{figure}[!htbp]
\centering
    \includegraphics[width=9cm,height = 2.5cm]{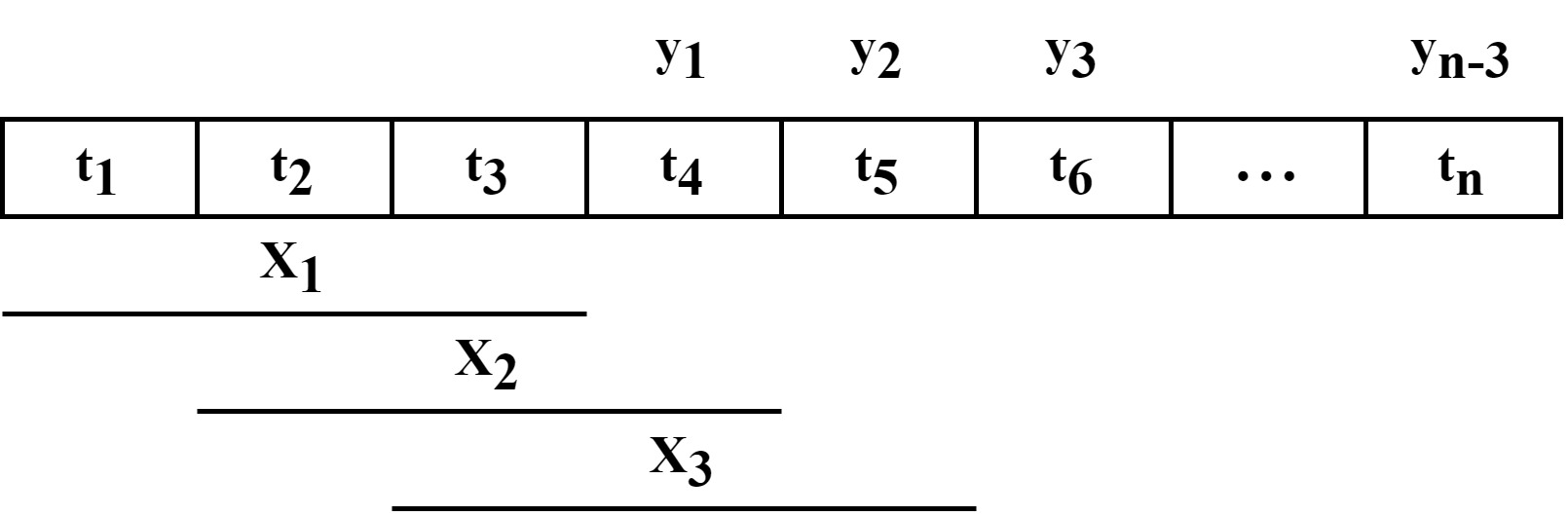}
    \caption{Data windowing}
    \label{fig:Data Windowing}
\end{figure}

Time series data need to be expressed into the proper format for supervised learning. Generally, the time-series data consists of several tuples (time, value), which is inappropriate for feeding them into the machine learning model. So, we restructured our original time series data using the sliding window technique. The sliding window technique is illustrated in Fig. \ref{fig:Data Windowing}. For example, three historical points ($X$) called features are used to predict the next data point ($y$) known as a target in the given sliding window example.
\subsection{Anomaly Detection and Mitigation}
\label{anomaly_detection}
In time-series data, the anomaly or outlier are those data points significantly different from the general trend or pattern of the entire data series. The statistical properties of the outliers are not aligned with the other data points in the time series.  There are mainly three ways to identify outliers in the data set: predictive confidence level approach, statistical profiling approaches, and clustering-based unsupervised approach. Our experiment used statistical profiling and an unsupervised learning model to identify the outlier in our time series data. We used the standard Three-Sigma rule to extract the inlier data points within three standard deviations from a mean of the time series. All other data points exceeding the boundary are considered outliers or anomalies.  We also applied an unsupervised machine learning model called Isolation Forest to compare the performance of two different types of outlier detection techniques. This technique follows a recursive algorithm to isolate the data points by randomly selecting an attribute and their splitting value between the minimum and maximum value of the attribute.

\subsection{Deep Learning Models}
\label{model}
\subsubsection{Recurrent Neural Network (RNN)}
RNN model is specifically designed to handle sequential data such as text mining, audio classification, language modeling, time series analysis, etc. The RNN uses the current sequence information and the previous sequence information to produce the current output at every step. Thus, the model learns information about all previous data points in the series at the last step. However, there is a short-term memory problem in the RNN model training process which is caused due to the vanishing gradient issues.



\subsubsection{Long Short-Term Memory (LSTM)} The purpose of the LSTM is similar to the GRU model. There are two additional gates called forget and output gate, along with the update and reset gate. LSTM has more control in transferring information among cells of the network. LSTM network is popular for processing time-series data to classify and make predictions. It alleviates the inherent vanishing gradient problem of the traditional RNN model and performs comparatively better.

\subsubsection{Gated Recurrent Unit (GRU)} The GRU has been proposed to solve the short-time memory problem in the RNN model. The concept of the gate is used in this model to control the flow of information between two consecutive cells. GRU model has an update gate that decides whether to transfer the previous cell output to the next cell or not. Gate is a mathematical unit that can measure the importance of the information and determine whether it should be stored or not. There are two gates called update gate and reset gate GRU model, which works on the update of cell state.

\subsubsection{LSTM Encoder-Decoder (LSTM\_En\_De)} This model can predict an output sequence from an input sequence known as the sequence-to-sequence model. It consists of two recurrent neural networks; one is called an encoder, and the other is a decoder. The encoder converts the input sequence into a fixed-length context vector and passes it to the decoder. The decoder uses the context vector and the final state of the encoder as the input and returns a sequence of output. 

\subsubsection{LSTM Encoder-Decoder with Attention Layer (LSTM\_EN\_DE\_Atn)} 
This model can also predict an output sequence from an input sequence, known as the sequence-to-sequence model. However, there is a drawback of the conventional encoder-decoder model, which is solved by adding an extra layer called attention layer, first proposed in \cite{vaswani2017attention}. The encoder-decoder model cannot extract the strong contextual relationship from long sequence data, which affects the model performance and decreases accuracy. On the other hand, the extra attention layer in the encoder-decoder model can identify the significance in sequence data.

\subsection{Time-series Cross-validation}
\label{cross_validation}
\begin{figure}[!htbp]
\centering
    \includegraphics[scale=0.5,width=9cm,height = 5cm]{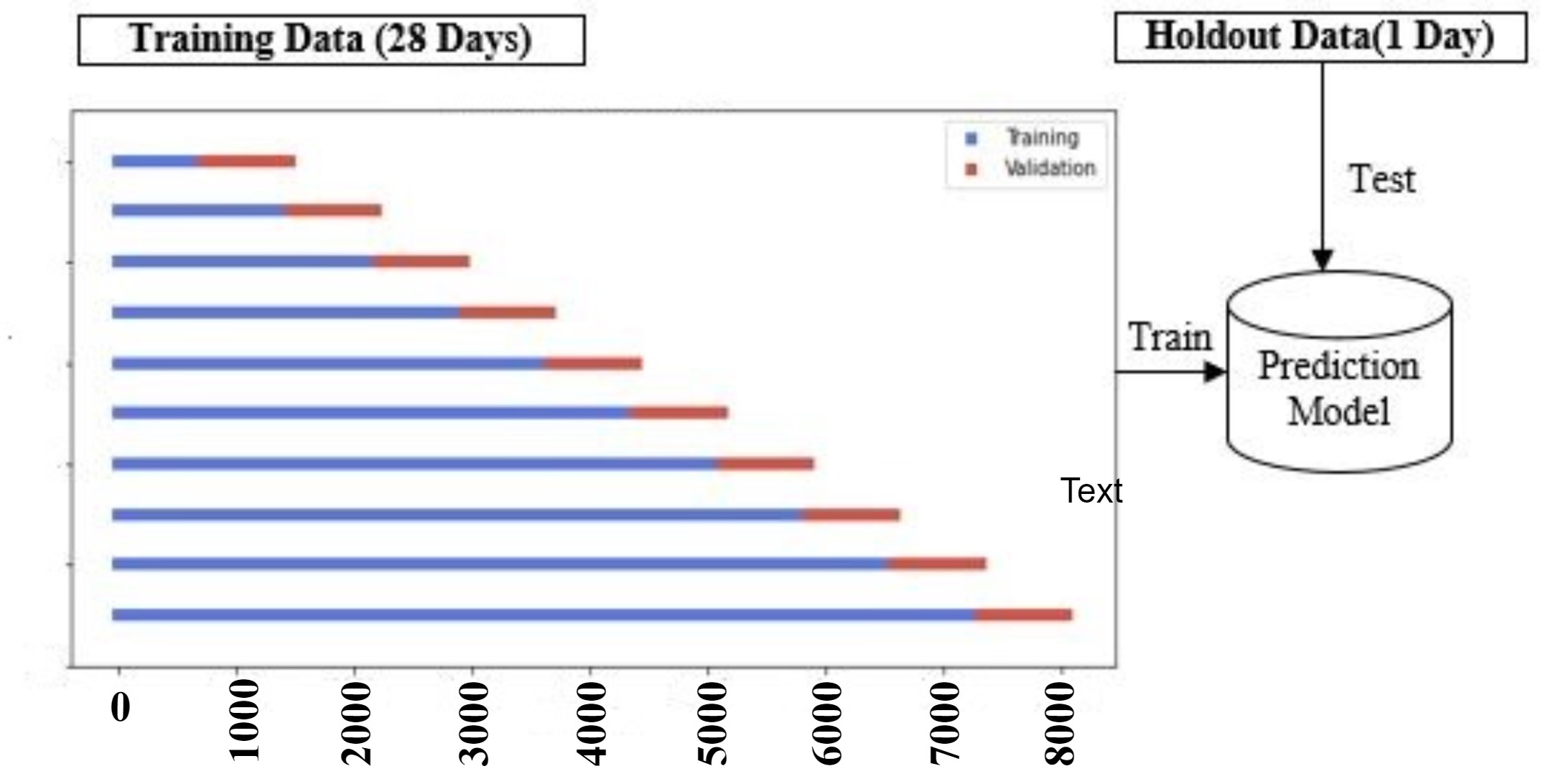}
    \caption{Time series cross-validation}
    \label{fig:cross validation}
\end{figure}

A cross-validation method was used in our experiment to evaluate the performance of our traffic prediction models. There are different ways of splitting the dataset into several folds to train and test the classification/prediction models. $K$-fold-cross validation splits the dataset into $K$ almost similar size of folds and all folds except one used to train the model while the remaining fold is kept for testing. The process continues until all the model is tested on every fold and the final performance of the model measured as the average performance on each fold. Since most of the cross-validation techniques in machine learning select fold randomly, we need to follow a different approach in splitting and selecting folds from time-series data to keep the temporal relation among folds. Our experiment used a rolling basis cross-validation technique where training starts with one-fold and finishes by predicting the next fold. In the next step, the test fold from the previous step is included in the training process and subsequent fold for the testing. The final performance of the model is the average of the prediction on each fold. We split our 29 days original time series data into two smaller datasets: training dataset of 21 days and holdout dataset of 8 days. The training dataset was cross-validated using the TimeSeriesSplit method from scikit-learn \cite{ scikit-learn} for model training, while the holdout dataset was used for testing. Fig. \ref{fig:cross validation} illustrates the cross-validation technique used in our experiment.

\subsection{Evaluation Metrics}
\label{metric}
We used Mean Absolute Percentage Error (MAPE) to estimate the performance of our traffic forecasting models. The performance metric identifies the deviation of the predicted result from the original data. For example, MAPE error represents the average percentage of fluctuation between the actual value and predicted value. Therefore, we can define our performance metric mathematically as follow: 

\begin{equation}
    MAPE = \dfrac{1}{n}\sum_{i=1}^n \bigg| \dfrac{p_i-o_i}{o_i} \bigg| \times 100 \%
\end{equation}
\begin{itemize}
    \item Here, $p_i$ and $o_i$ are predicted and original value respectively
    \item $n$ is the total number of test instance
\end{itemize}
\subsection{Software and Hardware Preliminaries}
\label{software}
We used Python and deep learning library TensorFlow-Keras\cite{chollet2015keras} to conduct the experiments.  Our computer has the configuration of Intel (R) i3-8130U CPU@2.20GHz, 8GB memory, and a 64-bit Windows operating system. 





\begin{figure}
    \includegraphics[width=9cm,height=5cm]{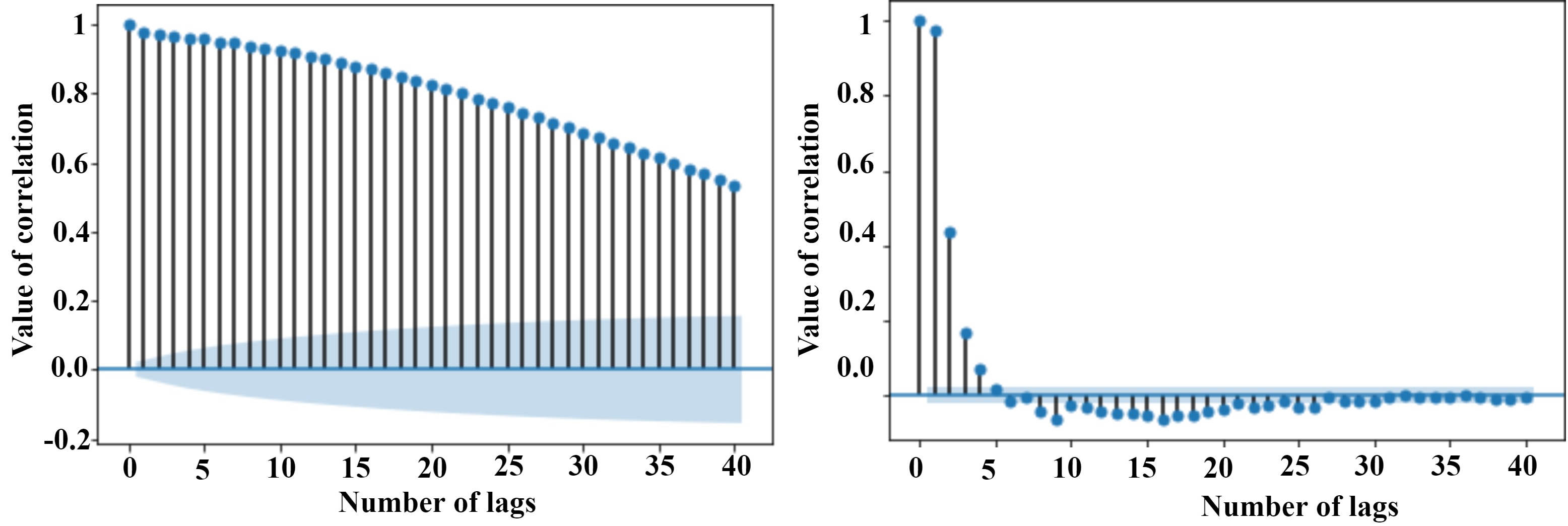}
    \caption{ACF and PACF Analysis}
    \label{fig:acf}
\end{figure}

\begin{figure}
\centering
        \includegraphics[width=9cm,height=4.1cm]{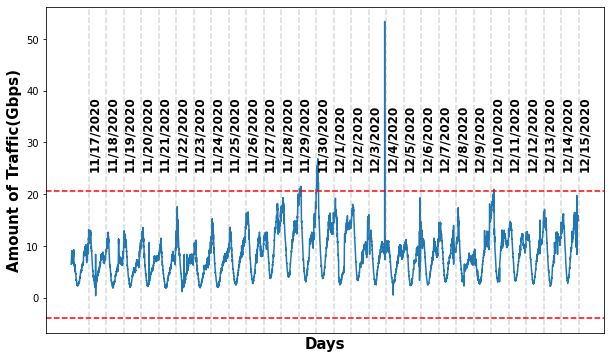}
        \caption{Outliers identified by Three-Sigma rule}
        \label{fig:three sigma}
\end{figure}%
\begin{figure}
        \centering
        \includegraphics[width=9cm,height=4cm]{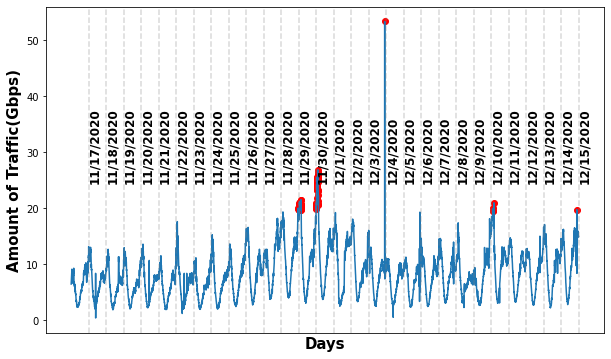}
        \caption{Outliers identified by Isolation Forest}
        \label{fig:isolation forest}
\end{figure}

\begin{figure*}[!htbp]
\centering
    \includegraphics[width=18cm,height = 7cm]{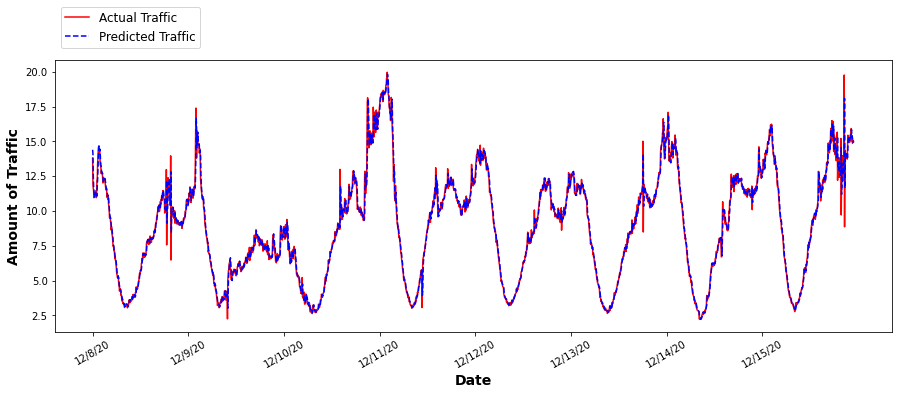}
    \caption{Actual traffic vs predicted traffic using LSTM\_En\_De model after adjusting outlier}
    \label{fig:Comparative result analysis}
\end{figure*}

\section{Results and Discussion}
\label{Results and Discussion}
Our experiment considered 21 days of IP traffic to train our deep learning models, while the last eight days of traffic were used for testing. Our experiment has three phases: I) traffic characteristics analysis, II) anomaly detection and mitigation, and III) deep sequence model implementation and performance evaluation. In phase I, several traffic characteristics, such as stationarity, seasonality, trend, etc., were analyzed to understand the traffic pattern. We analyzed our traffic data's ACF and PACF plots to determine the number of previous observations required to predict the next data points. For example, the PACF plot shows a significant value from lag 0 to lag 6, which indicates previous 6 data points can be important to predict the next one. Fig. \ref{fig:acf} [left] and Fig. \ref{fig:acf} [right] depicts the ACF and PACF plot, respectively. 
The real-world IP internet traffics is vulnerable to many external and internal factors. These factors unexpectedly change the normal traffic flow, which needs to be identified and alleviated for generalizing the learning capability by machine learning model. Two different methods were applied to determine the abnormal data points in our traffic in phase II. The Three-Sigma rule identifies those data points as outliers which are three standards that deviated from the mean of the traffic data. Fig. \ref{fig:three sigma} illustrates the inlier points wrapped by the upper and lower horizontal red line. The data points lie outside the top and bottom red line boundary in \ref{fig:three sigma} are identified as outliers according to the Three-Sigma rule.

\begin{table}
\caption{Performance summary for all model} 
\label{tab:Performance summary for all model}
\centering      
\begin{tabular}{l c c}  
\hline\hline                        
Model & MAPE (with outlier) & MAPE (without outlier)\\ [0.5ex] 
\hline                    
RNN & 7.51\% & 5.28\%   \\    
LSTM & 5.03 \% & 3.80 \%   \\ 
GRU & 6.41  \% & 5.28\%    \\ 
LSTM En\_De & 3.94\% &  3.51\%   \\ 
LSTM En\_De\_Atn & 3.95\% &  3.55\%   \\[1ex]       
\hline     
\end{tabular} 
\label{table:nonlin}  
\end{table}

Furthermore, an unsupervised learning algorithm called Isolation Forest is also used to identify the outliers, and the result is presented in Fig. \ref{fig:isolation forest}. From Fig. \ref{fig:three sigma} and Fig. \ref{fig:isolation forest}, a similar patterns of outliers are identified in our traffic data. We adjusted the outliers identified by the Three-Sigma rule before using them to train our prediction models. The technique used for adjusting the outliers is called backward filling, in which the following valid data point replaces the outlier.


Finally, in phase III, we applied several deep sequence models such as RNN and their variants, including LSTM, LSTM En\_De, LSTM En\_De\_Atn, and GRU to evaluate the comparative performance in traffic prediction.  All our model training continued for 100 epochs with a batch size of 16. We used five-fold cross-validation explained in section \ref {cross_validation} to train our models. The performance evaluation metrics of our experimental model are summarized in Table \ref{tab:Performance summary for all model}. Two different versions of each model were investigated to identify the impact of anomaly or outlier detection in traffic prediction. Our results show improved performance for prediction models trained on outlier-adjusted data. For example, the average deviation between actual and predicted traffic by RNN model is 7.51\% and 5.28\% with and without outlier, respectively, which improved traffic prediction by more than 29\% after adjusting outliers. In LSTM model, the average prediction error between actual and predicted traffic is reduced from 5.03\% to 3.80\%, i.e., more than 24\% after handling the outlier. Similarly, we noticed an error reduction of more than 11\% (3.94\% to 3.51\%) and 10\% (3.95\% to 3.55\%) due to the outlier adjustment in LSTM\_En\_De and LSTM\_En\_De\_Atn, respectively. The deviation between actual and predicted traffic is reduced by more than 19\% from 6.41\% to 5.28\% in GRU model. According to our experimental result, LSTM\_En\_De is the best prediction model with a minimum prediction error of 3.94\% with outliers in the data and 3.51\% without outlier. A graphical representation of real traffic and predicted traffic using the LSTM\_En\_De model after adjusting the outliers is shown in Fig. \ref{fig:Comparative result analysis}. Our experimental results show overall performance improvement for all considering deep sequence models after adjusting the abnormal traffic. However, in the real world, many internal and external factors can affect the regular traffic pattern. Since machine learning-based traffic prediction algorithms learn the general pattern in the dataset and predict accordingly, it is essential to handle the outliers before providing them to learn. Otherwise, there is a chance of learning from abnormal traffic patterns, which can affect the prediction result. Our experimental results also showed that outliers in the data make the model performance poor than the clean data.

\section{Conclusion}
\label{Conclusion}
In this work, we experimented with several deep sequence models in internet traffic prediction. A total of five different deep learning models such as RNN, LSTM,  LSTM\_En\_De, LSTM\_En\_De\_Atn and GRU were implemented to analyze our real-world traffic. We applied these models with and without outliers in traffic to demonstrate the impact of anomalies in traffic prediction. Our experimental results show a significant performance improvement for all models after mitigating the anomalies. In the real world, future traffic is often tied with different external and internal factors, which should be treated first before providing them into a machine learning model to predict. Otherwise, the model will be unable to generalize the actual traffic pattern, and predictions will not be effective. In the future, we would like to extend this work by adding the attention mechanism in both single-step and multi-step traffic prediction.
\bibliographystyle{IEEEtran}
\bibliography{conference_101719}

\vspace{12pt}

\end{document}